\documentclass{article}

\usepackage[dblblindworkshop, final]{neurips_2025}
\workshoptitle{The First Workshop on Generative and Protective AI for Content Creation}

\usepackage[utf8]{inputenc}   % allow utf-8 input
\usepackage[T1]{fontenc}      % use 8-bit T1 fonts
\usepackage{microtype}        % micro-typography

\usepackage{hyperref}         % hyperlinks
\usepackage{url}              % simple URL typesetting
\usepackage{amsmath, amssymb, amsfonts, mathtools, amsthm}
\usepackage{nicefrac}         % compact symbols for 1/2, etc.
\usepackage{breqn}            % automatic line-breaking of displayed maths

\usepackage{graphicx}
\usepackage{booktabs}         % professional-quality tables
\usepackage{subcaption}       % for subfigures
\usepackage{multirow}
\usepackage{multicol}

\usepackage{tikz}
\usetikzlibrary{positioning, shapes, arrows, calc, decorations.pathreplacing}

\usepackage{enumerate}
\usepackage{enumitem}
\usepackage{listings}
\usepackage{cleveref}
\crefname{section}{Sec.}{Secs.}
\crefname{appendix}{App.}{Apps.}
\crefname{result}{Res.}{Res.}
\crefname{equation}{Eq.}{Eqs.}
\crefname{definition}{Def.}{Defs.}
\crefname{figure}{Fig.}{Figs.}
\crefname{tabular}{Tab.}{Tabs.}
\crefname{table}{Tab.}{Tabs.}

\usepackage{amsmath,amsfonts,amssymb,bm,bbm,pifont,mathtools,mathdots}
\usepackage{thmtools, thm-restate}

\DeclareMathAlphabet{\mathsfit}{\encodingdefault}{\sfdefault}{m}{sl}
\SetMathAlphabet{\mathsfit}{bold}{\encodingdefault}{\sfdefault}{bx}{n}

\title{Dynamic VLM-Guided Negative Prompting for Diffusion Models}

\author{%
  Hoyeon Chang$^*$, Seungjin Kim$^*$, Yoonseok Choi\thanks{Equal contribution.}\\
  KAIST\\
  \texttt{retapurayo, sjkim, uooh77}@kaist.ac.kr
}

\begin{document}
\maketitle

% ------------------------------------------------------------------------------
\begin{abstract}
We propose a novel approach for dynamic negative prompting in diffusion models that leverages Vision-Language Models (VLMs) to adaptively generate negative prompts during the denoising process. Unlike traditional Negative Prompting methods that use fixed negative prompts, our method generates intermediate image predictions at specific denoising steps and queries a VLM to produce contextually appropriate negative prompts. We evaluate our approach on various benchmark datasets and demonstrate the trade-offs between negative guidance strength and text-image alignment.
\end{abstract}

% ------------------------------------------------------------------------------
\section{Introduction}
% Diffusion models \citep{sohl2015deep, ho2020denoising} have become a cornerstone of image generation, with Text-to-Image (T2I) models such as Stable Diffusion \citep{rombach2022high} seeing widespread use.
% Although text-to-image diffusion models are effective at generating images that closely match user prompts, the crucial task of filtering unwanted objects or concepts during generation remains challenging \citep{schramowski2023safe}.
% This is a significant issue, as growing ethical and privacy concerns surround the training and application of these models \citep{ZHANG2025102701, carlini2023extracting}.

A prevalent method for content filtering in T2I is negative prompting, which guides the model away from specified concepts through Classifier-Free Guidance (CFG) \citep{Ho2022ClassifierFreeDG}.
However, this technique suffers from key limitations.
First, it can disrupt the image generation process even when the unwanted content is not present, leading to over-correction and semantic drift \citep{yuanhao2024understanding, jinho2024contrastive, koulischer2024dynamic}.
Second, negative prompts are typically predefined, yet it is difficult to anticipate all potential unwanted elements that might arise from a given positive prompt \citep{Jaehong2024SAFREE}.
This can lead to either unnecessary or inefficient filtering.

In this work, we propose a novel solution: Vision-Language guided Dynamic Negative Prompting (VL-DNP).
By leveraging the advanced image and language understanding of modern open-source Vision-Language Models (VLMs), our method acts as a dynamic negative prompt generator.
VL-DNP detects the emergence of unwanted content during the denoising process and generates targeted negative prompts in real-time to erase it.
Importantly, our framework can be easily integrated into any pretrained CFG-based diffusion model without requiring joint training or model modifications.

% \textbf{Our contributions include:}
% \begin{itemize}
% \item A novel framework that integrates VLMs into the diffusion denoising process for adaptive negative prompt generation
% \item Comprehensive evaluation across multiple datasets demonstrating the impact of dynamic negative guidance
% \item Analysis of the trade-offs between content filtering strength and text-image semantic alignment
% \end{itemize}

% Overall, our work demonstrates that dynamic, context-aware negative prompting represents a promising direction for improving content safety in diffusion models while maintaining generation quality.

% --------------------------- Figure 1 -----------------------------------
\begin{figure}[t]
  \centering
  \includegraphics[width=\linewidth]{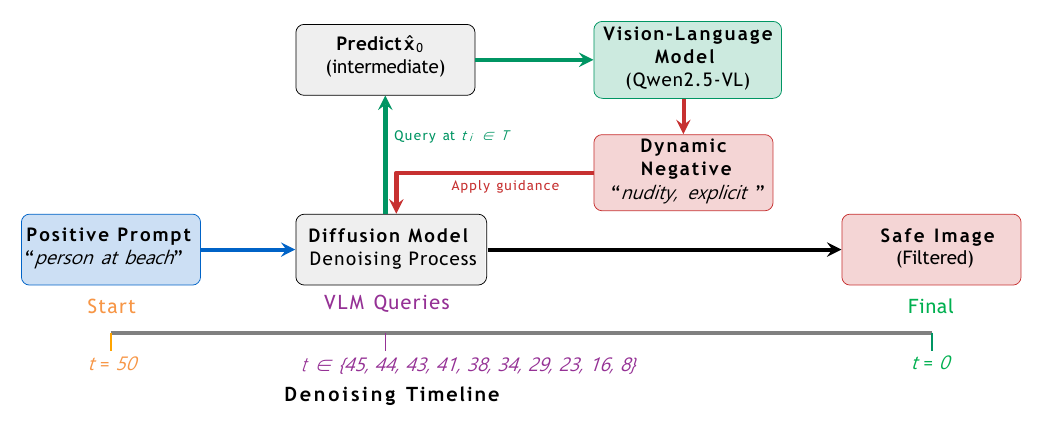}
  \caption{\textbf{VL-DNP inference pipeline.}  
           The positive text prompt is fed to a pretrained
           diffusion model.  At a small set of
           timesteps $t_i\!\in\!\mathcal{T}$ we predict the clean
           image $\hat{x}_0$, query a lightweight vision–language model
           (VLM) and obtain a \emph{dynamic negative prompt}.
           The prompt is fed back as classifier-free guidance, steering the
           remaining denoising steps away from any unsafe content detected in
           the intermediate image.}
  \label{fig:overview}
\end{figure}

% ------------------------------------------------------------------------------
\section{Related Work}
% \paragraph{Classifier-Free Guidance}

% Classifier Guidance \citep{dhariwal2021diffusion} is a method that uses a separate classifier $p_\theta(\bm{c}|\bm{x}_t)$ trained over the entire diffusion time step $t \in \mathcal{U}(0,1)$.
% Gradient of this classifier $\nabla_{\bm{x}_t}\log p(\bm{c}|\bm{x}_t)$ is used at inference time to guide our denoising trajectory to the distribution $\tilde{p}_0(\bm{x}_0) \propto {p}_0(\bm{x}_0){p}_0(\bm{c}|\bm{x}_0)^\omega$.
% However, due to the amount of training and the instability of the classifier, classifier-free guidance (CFG) was proposed \citep{Ho2022ClassifierFreeDG}.
% By replacing $\nabla_{\bm{x}_t}\log p(\bm{c}|\bm{x}_t)$ with $\nabla_{\bm{x}_t}\log p(\bm{x}_t|\bm{c})-\nabla_{\bm{x}_t}\log p(\bm{x}_t)$, we can obtain following augmented score function used to guide in classifier-free guidance.
% Classifier-Free Guidance (CFG) enhances diffusion model performance by combining conditional and unconditional score estimates \citep{Ho2022ClassifierFreeDG}. The standard CFG formulation uses a fixed guidance scale $\omega$ throughout generation:
% \begin{equation} 
%   {s}_{\theta,\text{cfg}}(x_t, t, c) \;=\;
%   \nabla_{{x}_t}\log p({x}_t) \;+\;
% \omega\Big(\nabla_{{x}_t}\log p({x}_t|{c})\;-\;\nabla_{{x}_t}\log p({x}_t)\Big)
%   \label{eq: cfg}
% \end{equation}

\paragraph{Negative prompting and Negative Guidance}

Negative prompting was proposed to filter out unwanted content.
In this work, we write positive conditions we like to align as $\textcolor{blue}{{c}+}$, and negative conditions we want to filter as $\textcolor{red}{{c}-}$.

\begin{equation}
    \label{eq:negprompt}
  {s}_\theta(x_t, t, \textcolor{blue}{{c}+},\textcolor{red}{{c}-}) \;=\;
  \nabla_{{x}_t}\log p({x}_t|\textcolor{blue}{{c}+}) \;+\;
\omega\left(\nabla_{{x}_t}\log p({x}_t|\textcolor{blue}{{c}+})\;-\;\nabla_{{x}_t}\log p({x}_t|\textcolor{red}{{c}-})\right).
\end{equation}

In Eq.~(\ref{eq:negprompt}), conventional negative prompting is considered a particular example that blends positive and negative guidance. The corresponding score function augmented with both positive and negative guidance is defined as follows:

\begin{equation}
\begin{aligned}
{s}_{\theta,\text{mixed}}(x_t, t, \textcolor{blue}{{c}+}, \textcolor{red}{{c}-})
=~
    \nabla_{x_t}\log p(x_t) + &\omega_{\text{pos}}\Big(\nabla_{x_t}\log p(x_t|\textcolor{blue}{{c}+}) - \nabla_{x_t}\log p(x_t)\Big) \\
    &- \omega_{\text{neg}}\Big(\nabla_{x_t}\log p(x_t|\textcolor{red}{{c}-}) - \nabla_{x_t}\log p(x_t)\Big).
\end{aligned}
\label{eq:pos_neg_guidance}
\end{equation}

% Similarly, in compositional guidance approaches \citep{compglide}, this type of score augmentation enables sampling from distributions $\tilde{p}_0(\bm{x}_0)\propto{p_0(\bm{x}_0|{\textcolor{blue}{\textcolor{blue}{\bm{c}+}}})^\omega / p_0(\bm{x}_0|\textcolor{red}{\textcolor{red}{\bm{c}-}})^\omega}$.
% However, the importance of negative guidance at each intermediate generation step might differ.
% Building upon this intuition, recent work \citep{koulischer2024dynamic} attempts to determine the weight of the negative guidance scale dynamically.
% Compared to this approach, we apply an adaptive negative guidance strategy that changes both the content and strength of negative prompting throughout the generation process.

% Similarly, in Composed Glide ~\citep{compglide}, they used this kind of score augmentation to sample from distribution $\tilde{p}_0(\bm{x}_0)\propto{p_0(\bm{x}_0|\{\textcolor{blue}{\bm{c}+}})^\omega / p_0(\bm{x}_0|\textcolor{red}{\bm{c}+})^\omega$.

Using Eq.~(\ref{eq:pos_neg_guidance}) for sampling implies sampling from $\tilde{p}({x}_0) \propto p({x}_0)\frac{p(\textcolor{blue}{{c}+}|{x}_0)^{\omega_{\text{pos}}}}{p(\textcolor{red}{{c}-}|{x}_0)^{\omega_{\text{neg}}}}$.

The importance of negative guidance at each intermediate generation step might differ.
Building upon this intuition, recent work \citep{koulischer2024dynamic} attempts to determine the weight of the negative guidance scale dynamically based on the estimated $p_t(\textcolor{red}{{c}-}|{x}_t)$. However, their approach still relies on a pre-defined negative prompt specified prior to inference.
% However, in order to estimate dynamic guidance scale, it is necessary to track $p_t(\textcolor{red}{\bm{c}-}|\bm{x}_t)$ from the beginning, using the score functions conditioned on $\textcolor{red}{\bm{c}-}$.
% Due to this, estimating $p_t(\textcolor{red}{\bm{c}-}|\bm{x}_t)$ for a diverse set of $\textcolor{red}{\bm{c}-} \in \gC -$ becomes computationally expensive in the context of dynamic negative guidance, as it involves multiple evaluations of the diffusion model.
% Rather than estimating $p_t(\textcolor{red}{\bm{c}-}|\bm{x}_t)$ from previous denoising steps as in prior approaches, we base our estimation on $\hat{\bm{x}}_0$, a strong predictor of the denoising direction.
 
\paragraph{Vision-Language Models in Generation}
VLMs have demonstrated remarkable capabilities in understanding visual content and generating descriptive text \citep{radford2021learning, li2022blip, li2023blip, liu2023visual}.
% Recent work has explored their integration into generative processes \citep{nichol2021glide, ramesh2022hierarchical}, primarily for post-hoc evaluation or one-time guidance rather than dynamic real-time adaptation during the denoising process.
In this work, we harness the image recognition and language generation capabilities of VLMs to provide dynamically adapting negative prompts during diffusion model inference.

% ------------------------------------------------------------------------------
\section{Methodology}

% \subsection{Framework Overview}
% Our Vision-Language Guided Dynamic Negative Prompting (VL-DNP) framework extends standard CFG-based diffusion models by introducing adaptive negative prompt generation during the denoising process. The key innovation lies in leveraging intermediate image predictions to guide the generation of contextually appropriate negative prompts through a Vision-Language Model (VLM).

The framework operates through the following steps: (1) Perform standard CFG denoising steps using the positive prompt. (2) At predefined timesteps, predict the denoised image $\hat{x}_0$ from current latent $x_t$. (3) Query a VLM to generate negative prompts based on the predicted image $\hat{x}_0$. (4) Apply negative guidance using the VLM-generated prompt for subsequent denoising steps.

\paragraph{Mathematical Formulation}
Based on the negative prompting formulation in \cref{eq:pos_neg_guidance}, we introduce temporal adaptivity to the negative conditioning. Let $\Theta$ denote a Vision-Language Model, and $\mathcal{T} = {t_1, t_2, \ldots, t_k} \subseteq [0, T]$ be a predefined set of timesteps where VLM queries occur.
At each query timestep $t_i \in \mathcal{T}$, we predict the denoised image denoted by $\hat{x}_0^{(i)}$ using the current estimate as follows:

\begin{equation}
\hat{x}_0^{(i)} = \frac{x_{t_i} + ({1-\bar{\alpha}_{t_i}}) \cdot s_{\theta,\text{cfg}}(x_{t_i}, t_i, \textcolor{blue}{{c}+})}{\sqrt{\bar{\alpha}_{t_i}}},
\label{eq:x0_prediction}
\end{equation}
where $\bar{\alpha}_{t_i}=\prod_{s=1}^{t_i}(1-\beta_{s})$, and $\beta_{s}$ is the forward process variance at timestep $s$ \citep{ho2020denoising}.
Classifier-free guidance (CFG) is defined as:

\begin{equation*}
  {s}_{\theta,\text{cfg}}(x_t, t, \textcolor{blue}{{c}+}) \;=\;
  \nabla_{{x}_t}\log p({x}_t) \;+\;
\omega\left(\nabla_{{x}_t}\log p({x}_t|\textcolor{blue}{{c}+})\;-\;\nabla_{{x}_t}\log p({x}_t)\right).
\end{equation*}

% \begin{equation}
% \hat{x}_0^{(i)} = \frac{x_{t_i} - \sqrt{1-\bar{\alpha}{t_i}} \cdot \epsilon\theta(x_{t_i}, t_i, \textcolor{blue}{\bm{c}+})}{\sqrt{\bar{\alpha}_{t_i}}}
% \label{eq:x0_prediction}
% \end{equation}

% where $\bar{\alpha}_{t_i}$ is the cumulative noise schedule parameter at timestep $t_i$.
The VLM then generates an adaptive negative prompt:

\begin{equation}
\textcolor{red}{{c}-}_{t_i} = \Theta(\hat{x}_0^{(i)}, \mathcal{D}),
\label{eq:vlm_generation}
\end{equation}

where $\mathcal{D}$ represents optional few-shot demonstration examples provided to the VLM. Although $\hat{x}_0^{(i)}$ is generally blurred during the initial denoising stage, we find that the VLM can detect objects even at early stages.

\paragraph{VLM Integration}
% We employ a VLM, utilizing few-shot prompting to generate contextually appropriate negative prompts. 
The VLM receives two inputs: (1) the intermediate image prediction $\hat{x}_0^{(i)}$, (2) demonstration examples that guide the model toward generating appropriate negative content descriptors.
Our prompting strategy instructs the VLM to identify potentially inappropriate or unwanted visual elements in the predicted image and generate concise negative prompts to suppress such content. 
% The few-shot examples provide the VLM with context about the types of content to detect and the format of desired negative prompts.
\paragraph{Dynamic Negative Guidance}
For timesteps $t_i < t < \text{min}(t_{i+1},T)$, we apply the augmented score function:

\begin{equation}
\begin{aligned}
\tilde{s}_\theta(x_t, t, \textcolor{blue}{{c}+}, \textcolor{red}{{c}-}_{t_i})
=~
    &\nabla_{x_t}\log p(x_t|\textcolor{blue}{{c}+}) + \omega_{\text{pos}}\Big(\nabla_{x_t}\log p(x_t|\textcolor{blue}{{c}+}) - \nabla_{x_t}\log p(x_t)\Big) \\
    &- \omega_{\text{neg}}\Big(\nabla_{x_t}\log p(x_t|\textcolor{red}{{c}-}_{t_i}) - \nabla_{x_t}\log p(x_t)\Big)
\end{aligned}
\label{eq:dynamic_guidance}
\end{equation}

% In the velocity parameterization used by modern diffusion models, this translates to:

% \begin{equation}
% v_f = v_{\text{uncond}} + \omega_{\text{pos}}(v_{\text{obj}} - v_{\text{uncond}}) - \omega_{\text{neg}}(v_{\text{neg}} - v_{\text{uncond}})
% \label{eq:velocity_field}
% \end{equation}

% where $v_{\text{neg}}$ is computed using the most recently generated negative prompt $\textcolor{red}{{c}-}_{t_i}$, and $\omega{\text{neg}}$ denotes the negative guidance scale.
The key advantage of this formulation is that $\textcolor{red}{{c}-}_{t_i}$ adapts to the evolving image content, allowing for more precise and contextual content filtering compared to static negative prompting approaches.

% ------------------------------------------------------------------------------
\section{Experimental Setup}
\paragraph{Datasets}
To evaluate the general performance of image generation, we use 100 prompts randomly sampled from COCO-30K \citep{lin2014microsoft}.
We compute a CLIP and FID score for the COCO-100 prompt set. To evaluate the filtering of unsafe images, we use Ring-a-Bell-16 \citep{tsai2023ring}, P4D \citep{zhi2024prompt}, and Unlean-diff \citep{yimeng2024togenerate} datasets, which consist of adversarial prompts designed to test content filtering in T2I tasks.

% \subsection{Implementation Details}
%  \textbf{Base Model}: Stable Diffusion v1.4 \citep{rombach2022high}
%  \textbf{VLM}: Qwen2.5-VL-7B-Instruct
% \textbf{Scheduler}: DPM-Solver++ with 50 inference steps
% \textbf{VLM Query Steps}: $\{5,10,20,30,40\}$ of 50 total steps
% \textbf{Evaluation}: CLIP scores + Nudenet Attack Success Rate and Toxic Rate.
\paragraph{Implementation Details}
We use Stable Diffusion v1.4 \citep{rombach2022high} as the base diffusion model and Qwen2.5-VL-7B-Instruct \citep{bai2025qwen2} as the Vision-Language Model for dynamic negative prompt generation. The denoising process employs DPM-Solver++  \citep{lu2022dpm} with 50 inference steps, where VLM queries are performed at timesteps ${45, 44, 43, 41, 38, 34, 29, 23, 16, 8}$ out of the total 50 steps. We evaluate our approach using CLIP-based metrics for text-image alignment, complemented by NudeNet ~\citep{notai2019nude} Attack Success Rate and Toxic Rate measurements to assess content filtering effectiveness.

% \subsection{Baseline Comparisons}
% We compare against:
% \begin{enumerate}[leftmargin=*,itemsep=0pt]
%   \item Standard CFG ($\omega_{\text{neg}} = 0.0$)
%   \item Fixed negative prompting with various negative guidance scales
%   \item Our dynamic VLM-guided approach with various negative guidance scales
% \end{enumerate}
\paragraph{Baseline Comparisons}
We compare our approach against baseline configurations: fixed negative prompting with various negative guidance scales, token embedding projection method SAFREE ~\citep{Jaehong2024SAFREE} and our dynamic VLM-guided approach across different negative guidance scale settings.

% ------------------------------------------------------------------------------
\section{Results}
\label{sec:results}

\begin{table}[t]
\centering
\caption{Safety (\textbf{ASR}, \textbf{TR}) vs.\ alignment/quality
         (\textbf{CLIP}, \textbf{FID}$\downarrow$).  
         “VL–DNP’’ = our dynamic VL-guided negative prompting,
         “Neg’’ = conventional static negative prompting.
         Lower is better for ASR, TR, and FID; higher is better for CLIP. InfT denotes the average time required to generate a single image. }
\label{tab:main_results}
\begin{tabular}{@{}lccccccccc@{}}
\toprule
 & \multicolumn{2}{c}{Ring-a-Bell-16} & \multicolumn{2}{c}{P4D} &
   \multicolumn{2}{c}{Unlearn-Diff} & \multicolumn{2}{c}{COCO-100} & \multirow{1}{*}{InfT} \\
\cmidrule(lr){2-3}\cmidrule(lr){4-5}\cmidrule(lr){6-7}\cmidrule(l){8-9}\cmidrule(l){10-10}
\textbf{Method} & ASR $\downarrow$ & TR $\downarrow$ & ASR & TR & ASR & TR & CLIP $\uparrow$ & FID $\downarrow$ & $ \text{sec} \downarrow$ \\ \midrule
SD\,v1.4 (no neg)      & 0.958 & 0.961 & 0.960 & 0.935 & 0.697 & 0.734 & 0.312 & --   & 8.0 \\ \midrule
\multicolumn{10}{l}{\textit{VL-DNP (Ours)}} \\
\quad $\omega_{\text{neg}}=7.5$  & 0.495 & 0.521 & 0.497 & 0.547 & 0.310 & 0.365 & 0.312 &  8.0  & 29.673 \\
\quad $\omega_{\text{neg}}=15.0$ & 0.084 & 0.147 & 0.225 & 0.277 & 0.099 & 0.171 & 0.311 & 12.9  & -- \\
\quad $\omega_{\text{neg}}=20.0$ & 0.011 & 0.081 & 0.113 & 0.163 & 0.085 & 0.139 & 0.311 & 15.3  & -- \\
\quad $\omega_{\text{neg}}=25.0$ & 0.032 & 0.068 & 0.086 & 0.134 & 0.077 & 0.130 & 0.311 & 15.1  & -- \\ \midrule
\multicolumn{10}{l}{\textit{Static negative prompting}} \\
\quad $\omega_{\text{neg}}=7.5$  & 0.200 & 0.295 & 0.298 & 0.373 & 0.204 & 0.252 & 0.313 & 107.3 & 11.967 \\
\quad $\omega_{\text{neg}}=15.0$ & 0.000 & 0.028 & 0.053 & 0.078 & 0.049 & 0.083 & 0.296 & 136.1 & -- \\
\quad $\omega_{\text{neg}}=20.0$ & 0.025 & 0.024 & 0.000 & 0.030 & 0.007 & 0.027 & 0.277 & 152.5 & -- \\ \midrule
SAFREE                           & 0.453 & 0.531 & 0.377 & 0.445 & 0.225 & 0.267 & 0.315 & 104.3 & 9.01 \\ \bottomrule
\end{tabular}
\end{table}

% --------------------------------------------------------------------------
Table~\ref{tab:main_results} collects the headline numbers. 
We report Attack-Success Rate (ASR) and Toxic Rate (TR) to quantify safety, and
CLIP score together with FID to quantify fidelity. 
% The three adversarial suites—\textit{Ring-a-Bell-16},
% \textit{P4D} and \textit{Unlearn-Diff}—stress safety, while the benign
% \textit{COCO-100} set checks everyday image quality.

% A clear tension emerges in every dataset: 
In every dataset, raising the negative-guidance scale
lowers ASR/TR but tends to raise FID.  
% The \emph{shape} of that curve differs
% sharply by method.  
With static negative prompting, increasing the scale from
$\omega_{\text{neg}}=7.5$ to~20 drags CLIP from~0.313 to~0.277 and drives FID from~107 to
153.  The same sweep under our VLM-guided schedule leaves CLIP essentially
constant (0.312\,$\rightarrow$\,0.311) and keeps FID in the 8–15 range while
still cutting ASR, especially at $\omega_{\text{neg}}=20$ and~25. SAFREE sits at the
opposite extreme of the trade-off.  It preserves the highest CLIP score in the
table (0.315 on COCO-100) but does so by accepting far poorer safety. 
% ASR
% remains above 0.45 on Ring-a-Bell-16 and 0.37 on P4D, more than an order of
% magnitude higher than VL-DNP at comparable fidelity.  In other words, SAFREE
% ``buys'' alignment at the cost of protection.

\begin{figure}[t]
  \centering
  \includegraphics[width=0.32\linewidth]{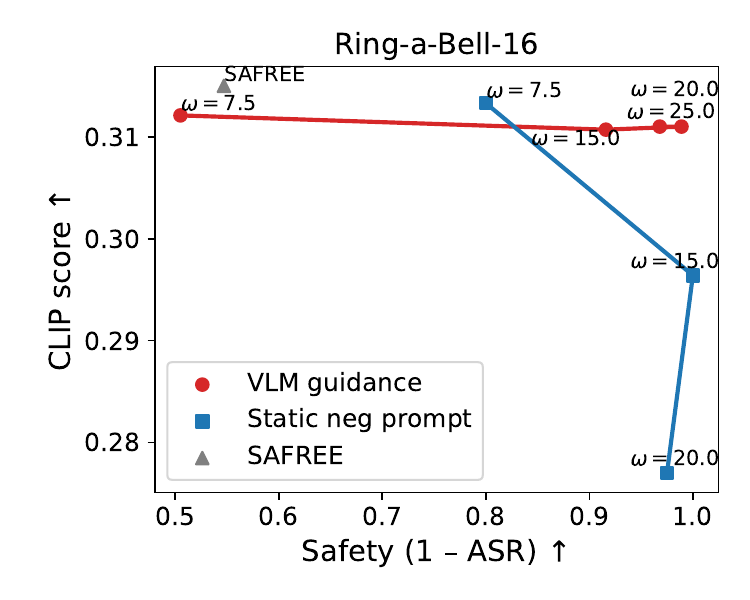}
  \includegraphics[width=0.32\linewidth]{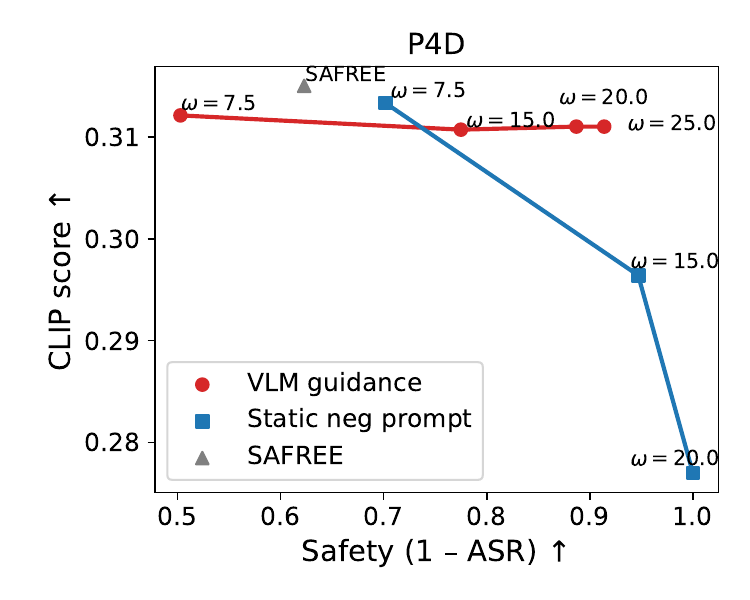}
  \includegraphics[width=0.32\linewidth]{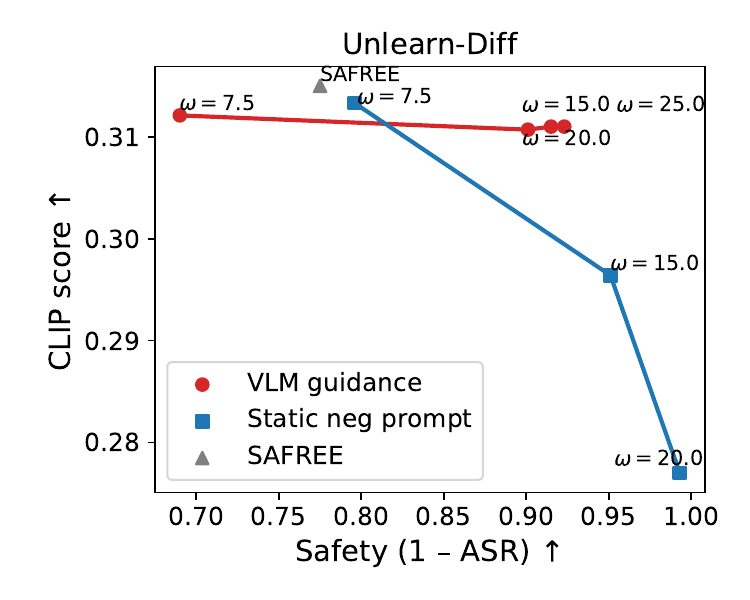}
  \caption{\textbf{Safety–alignment Pareto plots.}  Circles\,=\,dynamic
           \textbf{VLM-guided} prompts; squares\,=\,\textbf{static} prompts;
           triangles\,=\,\textbf{SAFREE} baseline.
           Axes follow ``larger\,=\,better'': Safety $(1{-}\text{ASR})$ on $x$,
           CLIP on~$y$ Markers are labelled by guidance scale~$\omega$.}
  \label{fig:pareto}
\end{figure}

These relations are visualised in Fig.~\ref{fig:pareto}.  Every static point at
$\omega_{\text{neg}}\in\{7.5,15,20\}$ is dominated by at least one VLM-guided point, and
SAFREE is likewise dominated: for any CLIP it attains, VL-DNP matches or beats
it on ASR and usually on FID as well.  The dynamic strategy therefore pushes
the entire safety–alignment frontier outward.

% Plotting these trade-offs (\cref{fig:pareto}) shows every static operating
% point at $\{\omega=7.5,15,20\}$ is \emph{dominated} by at least one VLM-guided
% point—i.e.\ VLM guidance offers either strictly lower ASR (higher Safety) at
% equal CLIP or strictly higher CLIP at equal ASR.  Consequently, VL-DNP traces a
% new Pareto frontier in the safety–alignment plane.

% \paragraph{Dataset variance.}
% Ring-a-Bell-16 is the hardest benchmark: even aggressive static guidance
% ($\omega\!=\!20$) leaves ASR~=~0.025, whereas VLM guidance cuts that to 0.011
% \emph{and} preserves CLIP (0.311 vs.\ 0.277).  P4D shows the same qualitative
% pattern, and Unlearn-Diff again yields a VLM-dominated frontier despite the
% static $\omega\!=\!20$ point reaching ASR\,=\,0.007 with a heavy CLIP penalty.

% --------------------------------------------------------------------------
\section{Discussion}
\label{sec:discussion}

\paragraph{Is VLM guidance safer?}
At a fixed $\omega$ the static method can achieve lower raw ASR
(e.g.\ 0.053 vs.\ 0.225 at $\omega=15$ on P4D), but only at the cost of a much
larger CLIP loss (–5.4\,\% vs.\ –0.5\,\% relative to no-neg).  From a
multi-objective viewpoint, VLM guidance gives a better \textit{safety–fidelity}
trade-off.

\paragraph{Why does dynamic prompting help?}
\emph{(i)} The VLM may propose narrow, concept-specific negatives
(“Male breast”, “Buttocks”) instead of generic “nsfw”, reducing
collateral suppression.  
\emph{(ii)} Prompts evolve: once an artefact disappears the VLM drops the
irrelevant negative and targets new risks, avoiding over-suppression.

\paragraph{Limitations \& future work} 
Although VLM guidance dominates the static baseline in Pareto terms, its
absolute ASR still hinges on the guidance scale.  Jointly scheduling
\emph{strength} and \emph{content} of guidance, while trimming the 
% 15–20\,\%
runtime overhead via lightweight vision encoders, remains a promising direction for future work. While the proposed dynamic prompting introduces additional latency due to VLM queries, this overhead may be mitigated by caching intermediate predictions or querying the VLM less frequently (e.g., every $k$ steps). Future work may also leverage lightweight VLMs or distillation-based approaches to enable real-time deployment.

% ------------------------------------------------------------------------------
% \section{Conclusion}

% VL-DNP integrates a Vision-Language Model into the diffusion denoising loop to generate dynamic negative prompts.
% Experiments show that it achieves adaptive content filtering while largely preserving text–image alignment, pushing the Pareto frontier outward on three safety benchmarks.
% Balancing safety and semantic fidelity, however, remains an open challenge and a fertile ground for future work.
% We presented VL-DNP, a novel approach for dynamic negative prompting in
% diffusion models using Vision-Language Models. Our evaluation highlights both
% the potential and limitations of this approach.

% \textbf{Key Achievements}
% \begin{itemize}[leftmargin=*]
%   \item Successfully integrated VLMs into the diffusion denoising process
%   \item Demonstrated adaptive content filtering capabilities
%   \item Provided empirical analysis of safety-quality trade-offs
% \end{itemize}

% \textbf{Future Directions}
% \begin{enumerate}[leftmargin=*,itemsep=0pt]
%   \item Develop metrics that better capture the safety-quality balance
%   \item Reduce overhead via VLM distillation or selective querying
%   \item Establish mathematical frameworks for dynamic negative guidance
%   \item Explore multi-modal guidance beyond vision-language models
% \end{enumerate}

% While VL-DNP represents a step toward more adaptive content filtering in
% diffusion models, balancing safety and semantic fidelity remains an open
% challenge.

% ------------------------------------------------------------------------------
\section*{Acknowledgements}
\label{sec: acknowledgement}

This work was supported by Institute of Information \& communications Technology Planning \& Evaluation(IITP) grant funded by the Korea government(MSIT) (No.RS-2020-II200940,Foundations of Safe Reinforcement Learning and Its Applications to Natural Language Processing).

\bibliographystyle{unsrtnat}
\bibliography{references}

\end{document}